\documentclass{article}

\usepackage{graphicx}
\usepackage{algorithm}
\usepackage{algpseudocode} 
\usepackage{color}
\usepackage{float,ulem,hyperref}
\usepackage{natbib}

\definecolor{pblue}{rgb}{0.13,0.13,1}
\definecolor{pgreen}{rgb}{0,0.5,0}
\definecolor{pred}{rgb}{0.9,0,0}
\definecolor{pgrey}{rgb}{0.46,0.45,0.48}

\usepackage{listings}
\definecolor{maroon}{rgb}{0.5,0,0}
\definecolor{darkgreen}{rgb}{0,0.5,0}
\lstdefinelanguage{XML}
{
  basicstyle=\ttfamily,
  morestring=[s]{"}{"},
  morestring=[s]{'}{'},
  morecomment=[s]{?}{?},
  morecomment=[s]{!--}{--},
  commentstyle=\color{darkgreen},
  moredelim=[s][\color{black}]{>}{<},
  moredelim=[s][\color{red}]{\ }{=},
  stringstyle=\color{blue},
  identifierstyle=\color{maroon}
}

\title{Synthetic Language Generation and Model Validation in BEAST2}
\date{\today}
\author{Stuart Bradley \\ Supervised by Dr. David Welch}

\newcommand{\dw}[1]{{\color{red} #1}} 
\newcommand{\code}[1]{\texttt{#1}}

\begin{document}

	\pagenumbering{gobble}
	\maketitle
    
    \begin{abstract}
	\noindent Generating synthetic languages aids in the testing and validation of future computational linguistic models and methods. 
This thesis extends the BEAST2 phylogenetic framework to add linguistic sequence generation under multiple models. The new plugin is 
then used to test the effects of the phenomena of word borrowing on the inference process under two widely used phylolinguistic models. 
    \end{abstract}
    
    \newpage
	\tableofcontents
    \addcontentsline{toc}{section}{Acknowledgments}
	\pagenumbering{arabic}
    
    \section*{Acknowledgments}
    
    If I begin by saying this thesis wasn't a serious trial, I would be doing both you (as the reader), and myself a major disservice. Despite my many trepidations towards doing postgraduate academic work, there are a sizable number of people that dragged me through honours --- at times kicking and screaming. It is these next few paragraphs that I dedicate to them, and all their help and hard work. 

Firstly and most importantly, I must thank my supervisor: Dr. David Welch. While his expertise and guidance were greatly appreciated, I am most thankful for his patience in dealing with not only a stubborn, but at times an ignorant student. Thank you for sticking with me, and helping me create something I am truly proud of. 

Next up I would like to thank members of the newly created Centre for Computational Evolution. Not only for allowing me to bounce around ideas and thoughts, but also for aiding me in both the theoretical and computational pursuits of this thesis. While the entire department was of great help, I'm going to use the following lines to list a number of instrumental people: Dr Remco Bouckaert, Dr Luke Maurits, Dr Walter Xie, William Hsu, Megan Ganley, and Arman Bilge. Thank you to you all.

Outside of the group I'd also like to give thanks to Prof Clark Thomborson and Dr Rizwan Asghar, who gave me advice on a previous paper, and helped me become far more confident in my academic communication. Additional thanks goes to Dr. Wayne Mitchell and Asela Dassanayake at Lanzatech for giving me the leeway required to focus on the thesis.

Finally, I come to my family (Jack, Alyson and Dana Bradley), my flatmates (Simon Corkindale, Kieran Newburgh, and Kathleen Seddon), and my large group of friends (with special reference to Erika Horlor). You have all been of immense help this past year, and I truly could not have done it without you. 

Once again, thank you all so much, and I hope you (the reader) find some level of enjoyment and education in the following body of work. 
    
    \newpage
    \section{Introduction}
    
Historical linguistics attempts reconstruct the complete history of language families in a systematic and sound way. Over the years there 
have been number of approaches to doing this, yet there does not appear to be a single perfect method. Without such a method, any 
suggested approach must be rigorously tested and validated before attempting to stand up to serious scrutiny. 
    
    \textit{Comparative linguistics} was the first such approach, helped in part by an explosion of linguistic data between 1785 and 1820 
\citep{Collinge-1995}. Compiled word lists were compared as proxies for entire languages, and could be used to determine the similarity of 
two languages (through \textit{cognate words}). While this method was by all intents systematic, it was often subjective, and lacked 
systematic attempts to validate the resulting hypotheses. 
    
    This subjectivity was partially quelled by the introduction of \textit{lexicostatistics} \citep{Swadesh-1955}. Swadesh developed a list of 
200 meanings that was considered universal among languages, known as a \textit{basic vocabulary}. By gathering word lists associated 
with each meaning for different languages, languages could be compared on a level playing field, and in theory, produced more accurate 
relationships, than could be determined when using earlier subjective judgments.
    
    \textit{Glottochronology} is an extension of the basic vocabulary concept, that includes word replacement at a constant rate, under a 
\textit{glottoclock}, related to the idea of a biological \textit{molecular clock}. From this, the percentages of shared cognates between 
languages can be converted into the length of time separating the languages in question. Larger percentages then suggest more recent 
separation. However, there are a number of issues that make glottochronology controversial \citep{Bergsland-1962}, such as the 
requirement in traditional glottochronology of a stable clock rate.  This has in turn lead to modified versions of glottochronology that 
attempt to fix earlier issues \citep{Kruskal-1973}. 
    
    In biology, \textit{phylogenetic inference} examines evolutionary  relationships between organisms, which produces a phylogeny. This is 
done via the comparison of heritable traits such as DNA or morphological traits. Phylogenetic inference also encompasses the use of 
computational techniques to  explore the space of probable trees--something that isn't touched upon in older comparative methods. 
    
    Phylogenetics also has the added benefit of pre-existing theory brought over from biology. When languages are represented by a string 
of absence or presence data for a number of traits, they can be treated in a similar manner to DNA sequences, so that very little 
modification is required to convert a biological phylogenetic method to a phylolinguistic one. 
    
    \newpage
    An example of this is in the selection of the best tree. As there are already a number of methods for finding biological evolutionary trees, 
the options for finding linguistic trees are also numerous \citep[see][for a recent review]{Sleator-2011}:
    \begin{enumerate}
		\item \textit{Distance-Matrix} - Sequence information is used to calculate distances between all pairs of organisms.  The 
resulting distance matrix is used to construct a tree. Trees with distances along branches between pairs of taxa which are closest to the 
distances in the matrix are considered the best. 
        \item \textit{Maximum-Parsimony} - The best tree is the one that requires the smallest number of evolutionary events to occur.
        \item \textit{Maximum-Likelihood} - Assigns probabilities to all possible phylogenetic trees via a model of substitution. The best tree is 
considered to be the one with the highest likelihood score. 
        \item \textit{Bayesian-Inference} - Prior probabilities, along with a likelihood function, are used to estimate the posterior density over 
possible phylogenies.  Rather than a single best tree, Bayesian methods typically produce a sampled set of trees representing a range of 
possible scenarios.  This is often done using Markov chain Monte Carlo sampling.
	\end{enumerate} 
    Initial attempts to enfold linguistics inside phylogenetic frameworks include work by \citep{Gray-2000}, \citep{Gray-2003}, and 
\citep{Atkinson-2005}. In all cases, the new inference method was used test already established hypotheses. These new methods brought 
some rigour to long-running debates, which lead  to testing new ideas within this framework. 
    
\cite{Atkinson-2005} used phylogenetic inference to look at the origin of the Indo-European family of languages. \cite{Bouckaert-2012} 
addressed the same problem with the explicit incorporation of geography and produced strong  evidence for Indo-European having its 
origin in the agricultural expansion out of Anatolia. 
    
    To make sometimes complex  phylogenetic tools more accessible, a number of packages were developed. BEAST 
\citep{Drummond-2012} (and later BEAST2 \citep{Bouckaert-2014}) were created. These programs were designed around using Markov 
chain Monte Carlo as the basis for Bayesian inference to construct phylogenies and test evolutionary hypotheses.
    
    While BEAST2 has a sequence generating capacity, it is ill-suited for models of language evolution; currently lacking for example, 
common models used in the field such as the stochastic-Dollo (SD) model. 
    
    This thesis aims to extend the language data simulation capacity of BEAST2, and to use simulations to test the robustness of common 
phylolinguistic estimation techniques to model misspecification, including various types of borrowing and missing data. In doing so, a 
language synthesis package (LangSeqGen) will be described that can be used quickly and effectively in future experiments. 
    
    \section{Models and Algorithms for Simulating Language Evolution}
    
Before diving directly into the simulation, both the language data, and the tree-like way in which  it evolves, need to be discussed; this 
then leads naturally onto the models used. 
    
    \subsection{Cognates}
	
	When one language diverges from another, both meanings and the forms of individual words can change. However, change is often 
an incredibly long process that takes place over many separate small changes \citep{Greenhil-2010}. Cognates are words in different 
languages which share a common ancestry, and are expertly defined as having similar meanings. In biological terms, cognate words are 
synonymous with homologous traits. An example of one such cognate might be: \textit{night} (English), \textit{nuit} (French), and 
\textit{noche} (Spanish), which all derive from the common root: *n\'{o}k\textsuperscript{w}ts (Proto-Indo-European). 
    
   While defining cognates is a systematic process, there are certain non-tree-like forces--such as borrowing--that muddy the process. To 
combat this, the Swadesh word list was developed. The list is a collection of 100-200 meanings that chosen to be culturally universal 
\citep{Swadesh-1955}.
	
	Given a common core of universal words, it then possible to compare different languages on the basis the presence or absence of 
specific cognates. However, direct comparisons--in the form of Glottochronology--rarely hold \citep{Bergsland-1962}.
	
	Instead of using cognate words for direct tree construction, they can be used as basis for defining a concise representation of a 
language, which can then be used in the models discussed below.    
	
	\subsection{Evolutionary Trees}

	Evolution is often represented as a tree \citep{Darwin-1859}. Starting with an ancestral species represented by a single lineage, 
evolutionary divergence causes the tree to branch. Splitting may be due to various circumstances, such as the separation between 
populations, which creates two distinct groups. Each lineage then continues to evolve separately through time causing the species to 
diverge. At the tips (leaves) of the tree, are the most modern--and possibly extinct (due to lack surviving information)--species. 
    \begin{figure}[H]
	\centering
			\includegraphics[width=8cm]{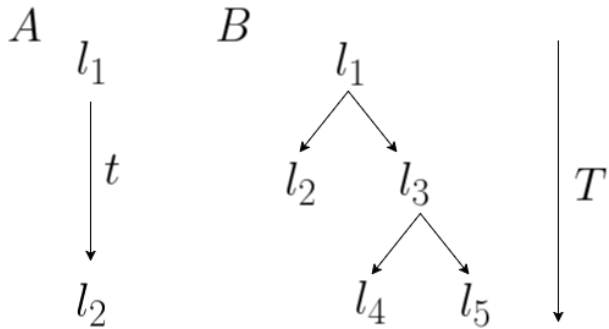}
			\caption{A: Single lineage evolution. B: Whole tree evolution.}
			\label{fig:exam_1}
		\end{figure}	
	Languages can also be represented in this fashion. In Figure \ref{fig:exam_1} A, language $l_{2}$ diverges from its ancestral 
language $l_{1}$; as the result of multiple accumulated evolutionary events, over time $t$. Over a larger time frame $T$, this event 
happens many times, to multiple languages, which results in a language evolution tree (see Figure \ref{fig:exam_1} B). In both cases, 
branch lengths can also vary, which is not shown in the figure.
	
	This tree begins with a root language, which branches so that languages can diverge as they accumulate different mutations. 
Computationally, this divergence is actually represented as an instant event. Instead of slow divergence, $l_{1}$ is copied to $l_{2}$, and 
both languages then mutate in accordance with some model of language evolution. This effectively demonstrates the idea that $l_{2}$ 
slowly moves away from its parent $l_{1}$.

	\subsection{Cognate Substitution Models}

	When modeling language evolution, languages need to be represented as abstract forms so they can be modelled and  processed 
efficiently. One approach to this is represent languages as the presence or absence of certain common cognate words, such as those in 
the Swadesh word list. For any language $l$, it can be represented as binary string. For example, if $l$ were being represented by the 
presence or absence of four cognates, it might look like:
	\begin{equation}
		\label{eq:lang_example}
		l = \left[\begin{array}{cccc} 0 & 1 & 1 & 0 \end{array}\right]
	\end{equation}
	Multiple languages can then be housed in a matrix (also known as an alignment), where each row represents a language and each 
column a cognate, to increase the compactness of the representation.  
	\begin{equation}
		\label{eq:C_string}
		C = \left[\begin{array}{cccc} 0 & 1 & 1 & 0 \\ 0 & 0 & 1 & 0 \\ 1 & 1 & 1 & 0 \end{array}\right]
	\end{equation}
	Once a representation is specified, the next step is to specify a model of language evolution. The evolution of a language happens 
as the result of mutations which in this case are the gain and loss of cognates.
	
	In the models discussed below, a site-independent \textit{Markov process} is used as the basis for evolutionary effects on languages. 
Site-independent means that it is assumed that mutations at any site are independent from any other site. The Markov assumption implies 
that times between mutations are exponentially distributed according to a rate parameter defined by the specific substitution model.
	
	\subsubsection{General Time-Reversible (GTR) Model}
	
	A large number of Markov models have been proposed to model DNA evolution, beginning with the Jukes\dw{-}Cantor model 
\citep{Jukes-1969}, which used equal mutation rates. To extend this model to allow for in-homogeneous rates, the GTR model 
\citep{Tavare-1986} was developed. Under this model, rates of transition between states can be variable, making it more suitable for 
language evolution. Under language evolution, the number of states is lower than biological evolution. Traits are represented by presence 
or absence, whereas in biological evolution there are 4 bases when modeling DNA or RNA, and 20 bases when modeling protein 
sequences. 
	The GTR model is a Markov process, consisting of a transition matrix, which is parameterised by time: \textit{t}. 
	
	Let
	\begin{equation}
		\label{eq:CTMC_Pt}
		P(t) = \left[\begin{array}{cc} p_{00}(t) & p_{01}(t) \\  p_{10}(t) & p_{11}(t)  \end{array}\right]
	\end{equation}
describe the probabilities of going from one state to another within \textit{t}. In the simplest case: $p_{01}=p_{10} = \mu$, the matrix is 
reduced to a single parameter. When applied to each site in \textit{l}, the result is \textit{l'}. To calculate \textit{P(t)}, it is possible to use a 
\textit{standard transition rate matrix} $Q$. 	
	\begin{equation}
		\label{eq:CTMC_Q}
		Q = \left[\begin{array}{cc} -\sum_{i\neq j} q_{ij} & q_{ij} \\ q_{ji} & -\sum_{i\neq j} q_{ij} \end{array}\right]
	\end{equation} 
	Matrix exponentiation can be used to derive $P(t)$ from $Q$ and \textit{t}:
	\begin{equation}
		\label{eq:P_t_exp}
		P(t) = \exp(Qt)
	\end{equation} 
	Given $P(t)$, it becomes a relatively simple programming exercise to iterate over the sites in \textit{l}, and at each site randomly pick 
a new state according to \textit{P(t)} as shown in Algorithm \ref{alg:GTR_1}. 
    
	\begin{algorithm}[H]
		\caption{GTR Simulation Algorithm}\label{alg:GTR_1}
		\begin{algorithmic}[1]
			\Function{GTR}{$l,Q,T$} \Comment{Mutate $l$ over time $T$ at rates $Q$}
				\State $P = MatrixExp(QT)$
				\For{$site$ \textbf{in} $l$}
					\State $site  = Rand(P)$ \Comment{Get a new site based on the transition probabilities in Equation 
\ref{eq:CTMC_Pt}}
				\EndFor
				\State \textbf{return} $l$
			\EndFunction
		\end{algorithmic}
        
	\end{algorithm}	
 What Algorithm \ref{alg:GTR_1} fails to capture is the individual evolutionary events, that is, the birth and death times of cognates. If this 
information is desired then the algorithm can be modified. Instead of using \textit{P(t)}, the \textit{Q} matrix is used directly, as follows:
 	\begin{algorithm}[H]
		\caption{GTR Simulation  Algorithm with explicit mutation events}\label{alg:GTR_2}
		\begin{algorithmic}[1]
			\Function{GTR}{$l,Q,T$} \Comment{Mutate $l$ over time $T$ at rates $Q$}
				\For{$site$ \textbf{in} $l$}
					\State $ t \sim Exp(site\:rate)$ 
					\While {$t < T$}
						\State $site  = Rand(Q)$ \Comment{Get a new site based on the transition rates in Equation 
\ref{eq:CTMC_Q}}
						\State $ t = t\:+\sim Exp(site\:rate)$ \Comment{Next mutation time, can be captured}
					\EndWhile
				\EndFor
				\State \textbf{return} $l$
			\EndFunction
		\end{algorithmic}
	\end{algorithm}
		
        \newpage
		\paragraph{Algorithm Comparison}
		\begin{figure}[H]
			\centering
			\includegraphics[width=12cm]{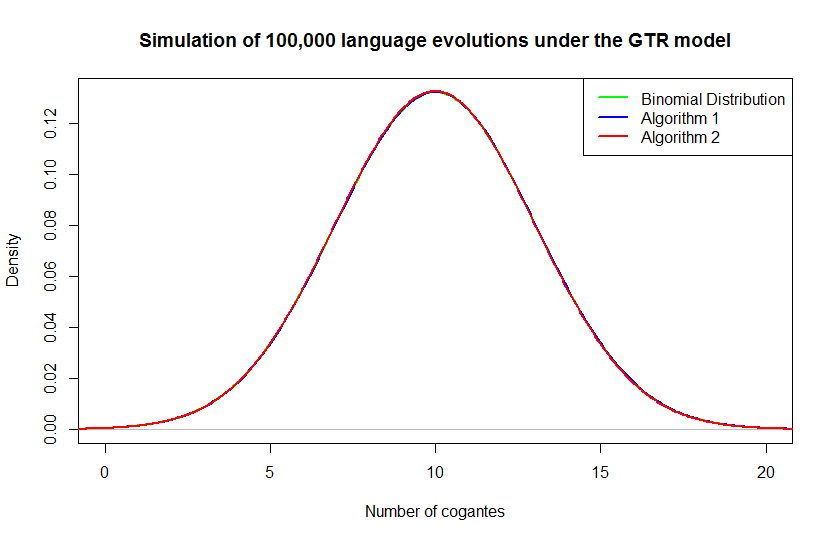}
			\caption{A comparison of the resulting languages from running Algorithms \ref{alg:GTR_1} and \ref{alg:GTR_2} 100,000 
times on random initial languages. The output from each algorithm was counted for the number of surviving cognates, which is plotted 
against the stationary binomial distribution.}
			\label{fig:GTR_sim}
		\end{figure}
		In Figure \ref{fig:GTR_sim}, a single language was independently mutated by  Algorithms 1 and 2  100,000 times each. The 
results of the simulation show that both methods produce the same distribution of the number of cognates. Parameters used in the figure 
are as follows: $Q = [[-0.5,0.5],[0.5,-0.5]]$, $T=100$, all starting languages were 20 traits in length. The simulations are plotted with the 
stationary binomial distribution, where $p$ is the probability of a $1$ given a sufficiently large $t$ -- see Equation \ref{eq:P_t_exp}. A 
binomial distribution is used due to the binary nature of the state space, traits are either absent or present. The final parameter, $n$ is the 
length of the language (20).
	\subsubsection{Covarion Model}
	The Covarion Model (where covarion is short for Concomitantly variable codons) is an extension of the standard GTR model. Instead 
of every site being variable, sites are able to move between variability and invariability at specific rates, $\delta$ and $\kappa$ 
\citep{Fitch-1970}. While this was initially developed to model site-specific changes of mutation rates, it can also be used to model 
cognates with differing rates of evolution over time, which is more consistent with linguists' view of language evolution 
\citep{Bouckaert-2012}.
	
    \newpage
	In the \textit{Q} matrix, each trait can now also move from a variant state to an invariant state, in addition to the states of trait birth 
and death. This means that the size of the initial matrix has to multiplied by a factor of four. 
	\begin{equation}
		\label{eq:covarion_matrix}
		 Q = \left[\begin{array}{cccc}
		 -(q_{01}+\delta) & q_{01} & \delta & 0 \\ 
		 q_{10} & -(q_{10}+\delta) & 0 & \delta\\ 
		 \kappa\delta & 0 & -(\kappa \delta) & 0 \\ 
		 0 & \kappa\delta & 0 & -(\kappa \delta)
		 \end{array}\right]
	\end{equation}
	\begin{equation}
		\label{eq:covarion_matrix_quad}
		Q = \left[\begin{array}{cc}
		A & B \\
		C & D
		\end{array}\right]
	\end{equation}
	The matrix in Equation \ref{eq:covarion_matrix} can be explained as the following four quadrants in Equation 
\ref{eq:covarion_matrix_quad}:
	\begin{enumerate}
		\item \textbf{A} - Original {\textit{Q} (variant) matrix}.
		\item \textbf{B} - Variant to invariant rates.
		\item \textbf{C} - Invariant to variant rates.
		\item \textbf{D} - Mutation rate in the invariant state.
	\end{enumerate}
	Equation \ref{eq:covarion_matrix} can then be applied to either Algorithms \ref{alg:GTR_1}, or \ref{alg:GTR_2}, without further 
modification. 
	\subsubsection{Stochastic-Dollo Model}
	Under the previous models, evolutionary change across a set of languages is considered \textit{time-reversible}. However, such an 
assumption is in direct contrast to the \textit{Dollo Parsimony} assumption, under which: ``An organism cannot return, even partially, to a 
previous state already realized in its ancestral series.'' \citep{Dollo-1893}. While many evolutionary features can be modeled using GTR it 
is argued that complex morphological traits are better suited to the Dollo Parsimony \citep{Gould-1970}. 
	
	The Dollo model is useful for language evolution because the creation of new cognates is itself a complex and rare event, making it 
extremely  unlikely the same word appears in two languages independently.
	
	Using the simplest form of the stochastic-Dollo model (binary state), no modification is required to move from biological to lexical 
evolution \citep{Nicholls-2006}. When evolving, languages gain \textit{unique} traits in a Poisson distributed fashion with rate $\lambda$. 
These traits are then lost at the \textit{per capita} Poisson rate $\mu$.

\newpage
This results in the following stationary distribution for the number of traits in a language \citep{Alekseyenko-2008}.
		\begin{equation}
			\label{eq:sd_ex}
			\frac{\lambda}{\mu}
		\end{equation}
		Evolutionary events are Poisson distributed along a branch; with an inhomogeneous rate that combines the rates of the birth 
and death Poisson processes, where $k$ is the number of traits in a language.
		\begin{equation}
			\label{eq:sd_rate}
			t \sim Exp(\lambda + k\mu),
		\end{equation} 
		Conditional on an event occurring, the probability that it is a birth event is:
		\begin{equation}
			\label{eq:sd_birthrate}
			\Pr(0 \rightarrow 1) = \frac{\lambda}{\lambda + k\mu}
		\end{equation}
        And the probability that it is a death event is:
		\begin{equation}
			\label{eq:sd_deathrate}
			\Pr(1 \rightarrow 0) = \frac{k\mu}{\lambda + k\mu}
		\end{equation}
		Once Equations \ref{eq:sd_rate}, \ref{eq:sd_birthrate}, and \ref{eq:sd_deathrate} have been defined for given values of $
\lambda$ and $\mu$, the process can be simulated using Algorithm \ref{alg:SD}.
        
	\begin{algorithm}[H]
		\caption{Stochastic-Dollo Simulation Algorithm}
		\label{alg:SD}
		\begin{algorithmic}[1]
			\Function{SD}{$l, \lambda, \mu,T$} \Comment{Mutate $l$ over time $T$ at rates $\lambda$, and $\mu$}
				\State $t \sim Exp(\lambda+k\mu)$ 
				\While {$t < T$}
					\State $u \sim U(0,1)$ 
					\If{$u \leq  \frac{\lambda}{\lambda + k\mu}$} \Comment{Birth event}
						\State $l\:+= new\:Trait$
					\Else \Comment{Death event}
						\State $l\:-= Random\:Trait$
					\EndIf			
					\State $t\:+\sim exp(\lambda+k\mu)$
				\EndWhile
			\EndFunction
		\end{algorithmic}
	\end{algorithm}
    
		\paragraph{Algorithm Validation}
		\begin{figure}[H]
			\centering
			\includegraphics[width=12cm]{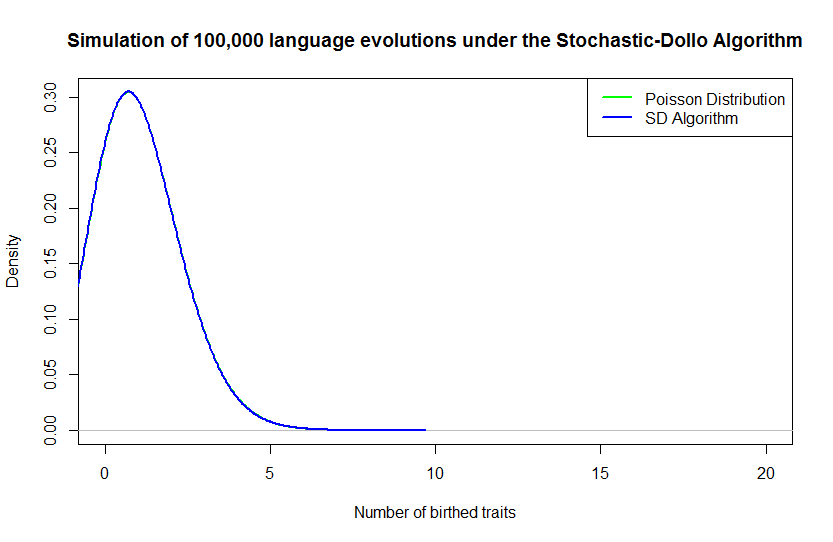}
			\caption{Result of running  Algorithm \ref{alg:SD} 100,000 times, and counting the number of birthed traits. This was then 
plotted against the stationary Poisson distribution, where the rate is defined by Equation \ref{eq:sd_ex}. Both $\lambda$ and $\mu$ are 
set at $0.5$, and $T = 10$.}
			\label{fig:SD_sim}
		\end{figure}
		The number of traits in the language output from Algorithm \ref{alg:SD} are Poisson distributed, with a mean derived from 
Equation \ref{eq:sd_ex}. Randomly sampling from the Poisson distribution can then be compared with the output of the algorithm to 
determine its correctness. Figure \ref{fig:SD_sim} illustrates that given sufficiently large $t$, any language will move towards the stationary 
distribution with a mean defined by Equation \ref{eq:sd_ex}. Language $l$ is fixed for all simulations, and is generated by giving it a 
number of traits drawn from the stationary distribution. 	
    
    \newpage
	\section{Whole Tree Sequence Generation}
    
    The next logical step from single lineage evolution, is to generate an entire tree from a single root language. 
    
    Once a tree structure is generated, it acts as a skeleton upon which languages are evolved. The time over which $l_{2}$ diverges from 
$l_{1}$ can be drawn from individual branch lengths.
        
        Whole tree generation is an extrapolation of single lineage evolution. Each individual branch evolves according to Algorithm 
\ref{alg:GTR_1}, \ref{alg:GTR_2}, or \ref{alg:SD}.
        
        \begin{algorithm}[H]
			\caption{Non-borrowing Tree Generation}
			\label{alg:whole_tree_gen}
			\begin{algorithmic}[1]
				\Function{Mutate Tree}{$tree, c$} \Comment {Mutate languages down $tree$ , and build Alignment $c$}
                	\State $currentParents = [tree.root]$
                    \State $newParents = []$
                    \While {$currentParents.length > 0$}
                    	\For {$parent$ in $currentParents$}
                        	\State $children = parent.getChildren$
                            \For {$child$ in $children$}
                            	\State $T = parent.height - child.height$ \Comment{Get branch length}
                                \State $l = mutate(parent.lang, T)$ \Comment{Mutate language}
                                \State $child.lang = l$
                                \State $c.add(l)$ \Comment{Add language to Alignment}
                                \State $newParents.add(child)$
                            \EndFor
                        \EndFor
                        \State $currParents = newParents$
                        \State $newParents = []$
                    \EndWhile
                    \State \textbf{Return} $tree, c$
				\EndFunction
			\end{algorithmic}
		\end{algorithm}
        
        Algorithm \ref{alg:whole_tree_gen} iterates through the nodes of $tree$, and mutates along its branches. Line 9 is where one of the 
previous algorithms returns a newly evolved language for the $child$ node, based on the language of its $parent$.
        
        	\subsection{Algorithm Validation}
           	\begin{figure}[H]
				\centering
				\includegraphics[width=12cm]{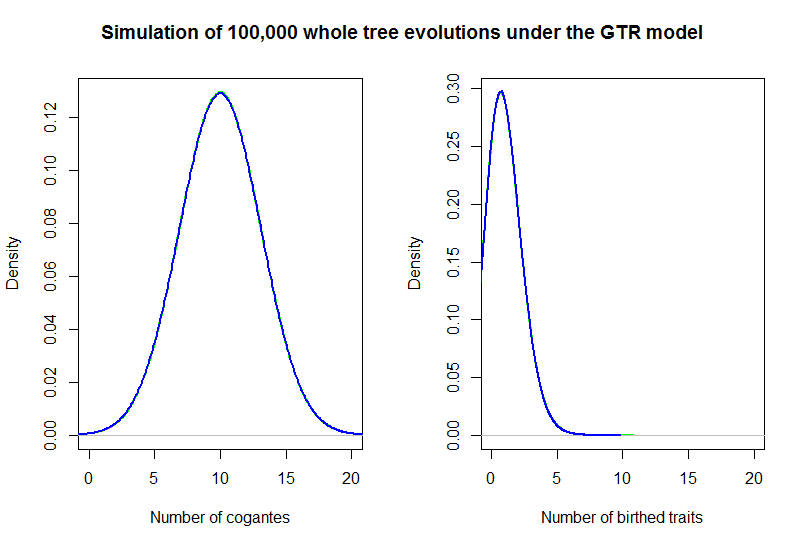}
				\caption{Comparisons of the different versions of Algorithm \ref{alg:whole_tree_gen} (blue), and their underlying 
distributions (green) as defined in Figures \ref{fig:GTR_sim}, and \ref{fig:SD_sim}. GTR is on the left, and SD is on the left.}
				\label{fig:Whole_tree_sim}
		\end{figure}
        Since Algorithm \ref{alg:whole_tree_gen} uses previously defined algorithms, over a whole tree, the leave nodes produce the same 
distribution of languages as with the individual algorithms. The parameters $l$ and $T$ can be redefined as the root language and tree 
height respectively. Figure \ref{fig:Whole_tree_sim} demonstrates that given for random trees with a high exponential branching rate 
($0.005$), these distributions hold true. 
        
        \newpage
        \section{Borrowing}
 
	In language evolution, borrowing is the phenomenon wherein a cognate from one language is borrowed by another language. This is 
analogous to horizontal gene transfer in biological evolution. More formally, borrowing events occur from a given language randomly at a 
per cognate exponential rate $b\mu$ . When a borrowing event occurs at time $t$, language $l_{2}$ receives a trait from $l_{1}$. Both the 
languages and traits are chosen at random, $l_{1}$ then gives a cognate to $l_{2}$, regardless of whether $l_{2}$ already has a trait or 
not. In the case where $l_{2}$ already has a trait, this can be considered replacement. 
	
	Borrowing runs concurrently with a cognate substitution model. In the case of GTR, borrowing is mathematically equivalent to a trait 
birth event, with a differing rate \citep{Atkinson-2005}. The stochastic-Dollo model complicates this slightly. Since individual traits are only 
born once, borrowing violates the Dollo assumption; when a trait is borrowed from $l_{1}$ to $l_2$, it can be considered to be born into 
$l_{2}$, meaning that an already dead trait can be born again. 
	
	To simulate borrowing, every evolutionary event must be captured, so as to accurately determine the number traits present in each 
language at any time, which determines the total rate of borrowing, and whether site $i$ has a trait at time $t$, from which borrowing can 
occur. This means that Algorithm \ref{alg:GTR_1} of the GTR model cannot be used (without modification), as it doesn't explicitly capture 
mutation events.
		\subsection{Global Borrowing Model}
		The global borrowing model (Algorithm \ref{alg:global_borrowing}), is the simplest way of modeling borrowing. Under this 
model, any language can borrow from any other language at $t$. 
        
		\begin{algorithm}[H]
			\caption{Global Borrowing Simulation Algorithm}
			\label{alg:global_borrowing}
			\begin{algorithmic}[1]
				\Function{GB}{$l_{1}, b, T$} \Comment{Borrow from $l_{1}$ over time $T$ at rate $b$}
					\State $t = T.height - (x \sim exp(b))$
					\While {$t > 0$}
						\State $L = \{x \: | \: x \: is \: alive \: at \: t \wedge x \neq l_{1} \} $ \Comment {Get all other languages $L$ 
alive at $t$}
						\State $l_{2} = Random(x \in L)$
						\State $site = Random ((site \in l \wedge site = 1))$
						\State $l_{2}(site) = 1$ \Comment {Birth randomly selected trait in $l_{2}$}
						\State $t\:+\sim exp(b)$
					\EndWhile
				\EndFunction
			\end{algorithmic}
		\end{algorithm}
		
        \newpage
		\subsection{Local Borrowing Model}
		Local borrowing deals with the concept that highly divergent languages are going to be less likely to borrow from one another 
\citep{Nicholls-2008}. Under local borrowing, $l_{2}$ can only borrow from $l_{1}$ if they share a common ancestor within time $z$. The 
algorithm for local borrowing is similar to Algorithm \ref{alg:global_borrowing}, and differs only in that it checks that $l_{1}$ and $l_{2}$ 
have a common ancestor within $z$. If there is not common ancestor, then the borrowing event does not occur.
        
		\begin{algorithm}[H]
			\caption{Local Borrowing Simulation Algorithm}
			\label{alg:local_borrowing}
			\begin{algorithmic}[1]
				\Function{LB}{$l_{1}, b, T, z$} \Comment{Borrow from $l_{1}$ over time $T$ at rate $b$, with distance $z$}
					\State $t  = T.height -(x \sim exp(b))$
					\While {$t > 0$}
						\State $L = \{x \: | \: x \: is \: alive \: at \: t \wedge x \neq l_{1} \} $ \Comment {Get all other languages $L$ 
alive at $t$}
						\State $l_{2} = Random(x \in L)$
						\State $site = Random ((site \in l \wedge site = 1))$
						\If {$dist(l_{1},l_{2}, z) == True$} \Comment {Check languages have common ancestor within $z$}
							\State $l_{2}(site) = 1$ \Comment {Birth randomly selected trait in $l_{2}$}
						\EndIf
						\State $t\:-\sim exp(b)$
				\EndWhile
				\EndFunction
			\end{algorithmic}
		\end{algorithm}
        
        Since the rate of language borrowing changes with the number of languages alive at time $t$ \citep{Atkinson-2005}, it is not possible 
to use the single lineage algorithms seen previously. Instead--for both GTR and SD--new total rates of evolution have to be defined with 
borrowing included, and from this new algorithms can be created and validated.
        
        \subsection{Adding Borrowing to Cognate Substitution Models}
        
        	\subsubsection{GTR Borrowing}
            
            Under the GTR model, mutations happen at a per site rate $\mu$, which can be represented by multiplying the mutation rate by 
the number of sites $\left |l\right |$, in a language $l$.
            \begin{equation}
				\label{eq:gtr_borrow_1}
				Mutation\:Rate = \mu \left |l\right |
			\end{equation} 
            Borrowing happens at a per alive (1) site rate $b\mu$, which can be represented similarly to Equation \ref{eq:gtr_borrow_1}; where 
the borrowing rate is multiplied by the number of alive sites $k$. 
            \begin{equation}
				\label{eq:gtr_borrow_2}
				Borrowing\:Rate = \mu b k_{l}
			\end{equation} 
            Equations \ref{eq:gtr_borrow_1}, and \ref{eq:gtr_borrow_2} can be combined and extended to $n$ languages to produce a total 
rate for any point on the tree.
            \begin{equation}
				\label{eq:gtr_total_rate}
				Total\:Rate = \mu \left( \sum_{i=1}^{n} \left |l_{i}\right | \right) + b \mu \left( \sum_{i=1}^{n}k_{i} \right)
			\end{equation} 
            Once the total rate is defined, the individual probabilities can be defined in a similar vein as the original stochastic-Dollo model. It 
then becomes possible to construct an algorithm to traverse and populate a tree with generated languages.
            
            \begin{algorithm}[H]
				\caption{GTR tree generation with borrowing}
				\label{alg:whole_tree_gen_gtr}
				\begin{algorithmic}[1]
					\Function{Mutate Tree}{$tree, borrow, z$} \Comment {Mutate $tree$, using $borrow$ rate, and $z$ for local 
borrowing distance} 
						\State $aliveNodes = \{x \: | \: x \: is \: alive \: at \: 0.0\} $ \Comment{Get alive languages at 0.0 (root)}
                		\State $totalRate = getTotalRate(aliveNodes)$
                		\State $t = tree.height - (x \sim exp(totalRate))$
                        \For {$event$ in $branchTimes$}
                        	\State $aliveNodes = \{x \: | \: x \: is \: alive \: at \: event\} $ 
                            \State $totalRate = getTotalRate(aliveNodes)$
                		\While {$t > event$}
							\State $u \sim U(0,1)$ 
                        	\If {$ u < \frac{Mutation\:Rate}{Total\:Rate}$} \Comment{Mutation event}
                        		\State $node = Random(aliveNodes)$ 
                            	\State $site = Random(node.language)$ \Comment{Pick random site in language}
                            	\State  $node.language[site] = 1 - site$ \Comment{$0 \rightarrow 1$ or $1 \rightarrow 0$}
                        	\Else \Comment{Borrowing event}
                        		\State $n_{1}, n_{2} = Random(x \in aliveNodes \wedge n_{1} \neq n_{2})$ \Comment {$n_1$ and $n_2$ are picked 
randomly with probability $n_{*} / totalBirths(aliveNodes)$}
								\State $site = Random ((site \in n_{1}.language \wedge site = 1))$
                        		\If {$dist(n_{1},n_{2}, z) == True$} \Comment {Check languages have common ancestor within $z$}
									\State $n_{2}.language[site] = 1$ \Comment {Birth randomly selected trait in $l_{2}$}
								\EndIf
                        	\EndIf
                            \State $totalRate = getTotalRate(aliveNodes)$
                        	\State $t\:-\sim exp(TotalRate)$
                		\EndWhile
                        \EndFor
					\EndFunction
				\end{algorithmic}
			\end{algorithm}
            
            \newpage
            \paragraph{Algorithm Validation} To confirm that Algorithm \ref{alg:whole_tree_gen_gtr} works, it can be compared to the 
multinomial distribution which underlies the interactions between multiple languages. 
                
                For two languages, each event in the tree moves the twin cognates from each language into a new state; this is defined by the 
following rate matrix.
                \begin{equation}
					\label{eq:gtr_borrow_rate_2}
                    Q=
					\begin{array}{ccccc}
                    	            & \mathbf{00} & \mathbf{01} & \mathbf{10} & \mathbf{11} \\ 
                        \mathbf{00} & -\sum_{i\neq j} q_{ij} & \mu & \mu & 0 \\ 
                        \mathbf{01} & \mu & -\sum_{i\neq j} q_{ij} & 0 & (b+1)\mu \\ 
                        \mathbf{10} & \mu & 0 & -\sum_{i\neq j} q_{ij} & (b+1)\mu \\ 
                        \mathbf{11} & 0 & \mu & \mu & -\sum_{i\neq j} q_{ij}
                    \end{array}
				\end{equation}
        		Then given predefined values for the rates ($\mu,b =0.5$), the stationary distribution can be determined by plugging in 
Equation \ref{eq:gtr_borrow_rate_2} into Equation \ref{eq:P_t_exp}, where $t$ tends towards infinity. 
                \begin{equation}
					\label{eq:gtr_borrow_rate_2_stationary}
                    \begin{array}{cccc}
						\mathbf{00} & \mathbf{01} & \mathbf{10} & \mathbf{11} \\ 
                        0.22\dot{2} & 0.22\dot{2} & 0.22\dot{2} & 0.33\dot{3} 
					\end{array}
				\end{equation}
                Equation \ref{eq:gtr_borrow_rate_2_stationary} then defines the probabilities for the underlying multinomial distribution. This can 
then be compared with the distribution of states output by a 2-leaf tree run through Algorithm \ref{alg:whole_tree_gen_gtr}.
                \begin{figure}[H]
					\centering
					\includegraphics[width=12cm]{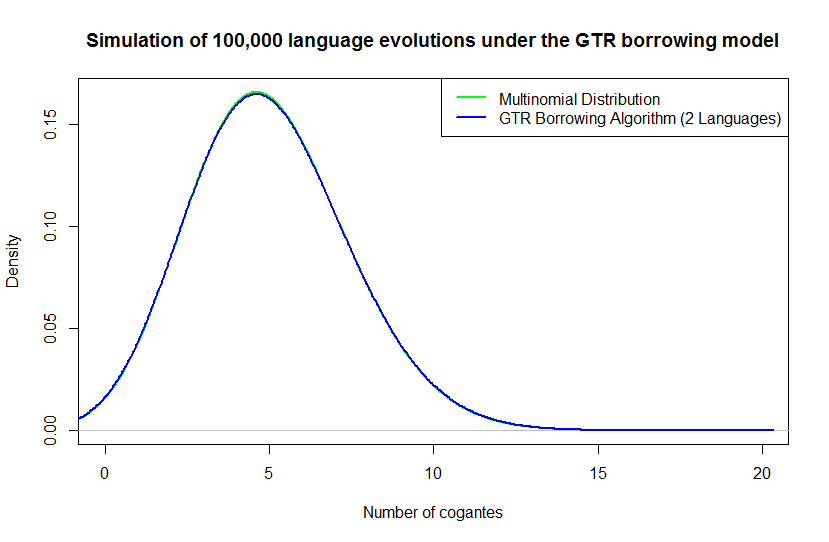}
					\caption{Comparison of theoretical multinomial distribution with an experimental multinomial distribution drawn 
from Algorithm \ref{alg:whole_tree_gen_gtr} using two languages.}
					\label{fig:gtr_borrow_multinom_2}
				\end{figure}
                In Figure \ref{fig:gtr_borrow_multinom_2}, there is a slight discrepancy between the two distributions. This is because the 
stationary distribution assumes languages of very high length; something that is not computationally tractable for 100,000 simulations. 
Instead, languages were 20 cognates in length.  
                To prove the result holds for a higher number of languages, larger matrices can be generated--from which stationary 
distributions can be calculated and compared. 
                
                The following calculations are for three languages. 
                \begin{figure}[H]
					\centering
					\includegraphics[width=13cm]{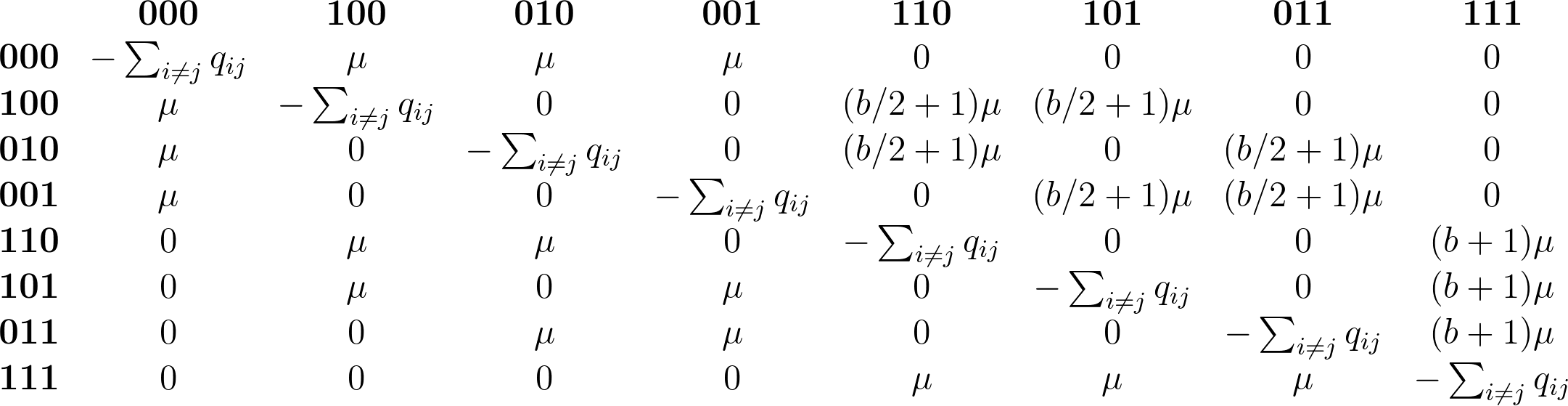}
					\label{fig:eq_17}
				\end{figure}
                \begin{equation}
					\label{eq:gtr_borrow_rate_3_stationary}
                    \begin{array}{cccccccc}
						\mathbf{000} & \mathbf{100} & \mathbf{010} & \mathbf{001} & \mathbf{110} & \mathbf{101} & \mathbf{011} 
& \mathbf{111} \\ 
                        0.0930 & 0.0930 & 0.0930 & 0.0930 & 0.1395 & 0.1395 & 0.1395 & 0.2093
					\end{array}
				\end{equation}
                \newpage
                \begin{figure}[H]
					\centering
					\includegraphics[width=12cm]{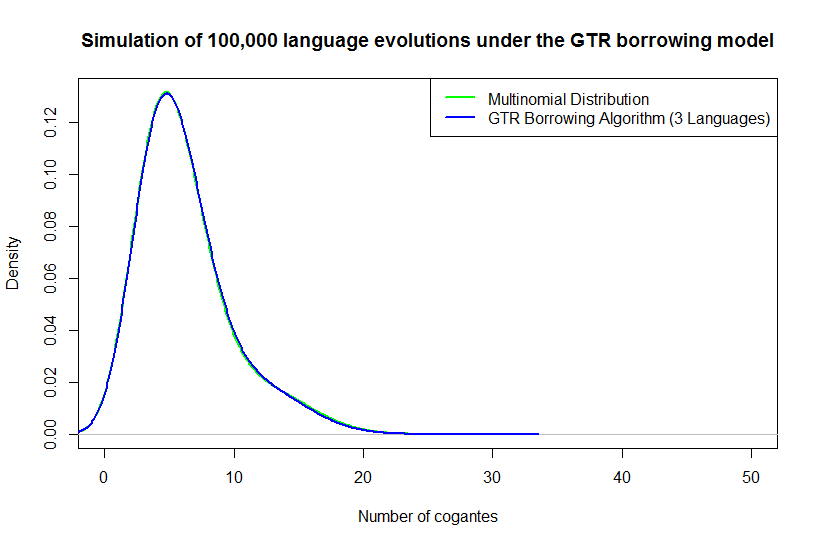}
					\caption{Comparison of theoretical multinomial distribution with an experimental multinomial distribution drawn 
from Algorithm \ref{alg:whole_tree_gen_gtr} using three languages.}
					\label{fig:gtr_borrow_multinom_3}
				\end{figure}
                Figure \ref{fig:gtr_borrow_multinom_3} once again demonstrates the discrepancy between experimental data, and the 
theoretical distribution defined by Equation \ref{eq:gtr_borrow_rate_3_stationary}. However, it does demonstrate that the algorithm is 
effective for multiple languages. 
                
            \subsubsection{Stochastic-Dollo Borrowing}
            
            Having both a birth rate --- $\lambda$ --- and a death rate --- $\mu$ -- complicates the stochastic-Dollo model somewhat when 
borrowing is introduced. The birth rate is tied to the number of languages alive, $\left |\bar{l}\right |$, at time $t$.
            \begin{equation}
				\label{eq:sd_birth_rate}
				Birth\:Rate = \lambda \left |\bar{l}\right |
			\end{equation}
            The death rate is proportional to the number of alive cognates $k$ in each language $i \in \left |\bar{l}\right |$.
            \begin{equation}
				\label{eq:sd_death_rate}
				Death\:Rate = \sum_{i=1}^{n}\mu k_{i}
			\end{equation}
            Finally, the borrowing rate is related to both the death rate, and the number of alive cognates in each language.
        	\begin{equation}
				\label{eq:sd_borrow_rate}
				Borrow\:Rate = \mu b \sum_{i = 1}^{n}k_{i}
			\end{equation}
            Combining Equations \ref{eq:sd_birth_rate}, \ref{eq:sd_death_rate}, and \ref{eq:sd_borrow_rate} produces the total rate of 
evolution across the tree.
            \begin{equation}
				\label{eq:sd_total_rate}
				Total\:Rate = \lambda \left |\bar{l}\right |+ \left ( \sum_{i=1}^{n}\mu k_{i} \right ) + \left (\mu b \sum_{i = 1}^{n}k_{i} 
\right )
			\end{equation}
            The algorithm for the stochastic-Dollo borrowing model is very similar to Algorithm \ref{alg:whole_tree_gen}, with the addition of 
separate birth and death events. 
            \begin{algorithm}[H]
				\caption{SD tree generation with borrowing}
				\label{alg:whole_tree_gen_sd}
				\begin{algorithmic}[1]
					\Function{Mutate Tree}{$tree, c, borrow, z$} \Comment {Mutate $tree$, and build Cognate Set $c$, using 
$borrow$ rate, and $z$ for local borrowing} 
						\State $aliveNodes = \{x \: | \: x \: is \: alive \: at \: 0.0\} $ \Comment{Get alive languages at 0.0 (root)}
                		\State $totalRate = getTotalRate(aliveNodes)$
                		\State $t = tree.height - (x \sim exp(totalRate))$
                        \For {$event$ in $branchTimes$}
                        \State $aliveNodes = \{x \: | \: x \: is \: alive \: at \: event\} $ 
                        \State $totalRate = getTotalRate(aliveNodes)$
                		\While {$t > event$}
							\State $u \sim U(0,1)$ 
                        	\If {$ u < \frac{Birth\:Rate}{Total\:Rate}$} \Comment{Birth event}
                        		\State $node = Random(aliveNodes)$ 
                            	\State  $node.language += 1$ \Comment{Add a cognate}
                            \ElsIf {$u > \frac{Birth\:Rate}{Total\:Rate}$ and $u < \frac{Birth\:Rate}{Total\:Rate} + \frac{Death\:Rate}{Total\:Rate}$} 
\Comment{Death event}

\State $node = Random(aliveNodes)$ 
                                \State $site = Random ((site \in n_{1}.language \wedge site = 1))$
                                \State $n_{2}.language[site] = 0$ \Comment {Kill randomly selected trait in $node$}
                        	\Else \Comment{Borrowing event}
                        		\State $n_{1}, n_{2} = Random(x \in aliveNodes \wedge n_{1} \neq n_{2})$ \Comment {$n_1$ and $n_2$ are picked 
randomly with probability $n_{*} / totalBirths(aliveNodes)$}
								\State $site = Random ((site \in n_{1}.language \wedge site = 1))$
                        		\If {$dist(n_{1},n_{2}, z) == True$} \Comment {Check languages have common ancestor within $z$}
									\State $n_{2}.language[site] = 1$ \Comment {Birth randomly selected trait in $l_{2}$}
								\EndIf
                        	\EndIf
    						\State $totalRate = getTotalRate(aliveNodes)$
                        	\State $t\:+\sim exp(TotalRate)$
                		\EndWhile
                        \EndFor
					\EndFunction
				\end{algorithmic}
			\end{algorithm}
            
            \newpage
            \paragraph{Algorithm Validation} Instead of determining the stationary distribution of three interacting Poisson processes, the 
algorithm can be validated against previous work. TraitLab \citep{Nicholls-2011} was created to perform some of the operations described 
in this thesis. Data synthesis--under the stochastic Dollo model--inside TraitLab produces sequences which can then be compared to 
similar output from Algorithm \ref{alg:whole_tree_gen_sd}. 
             \begin{figure}[H]
				\centering
				\includegraphics[width=12cm]{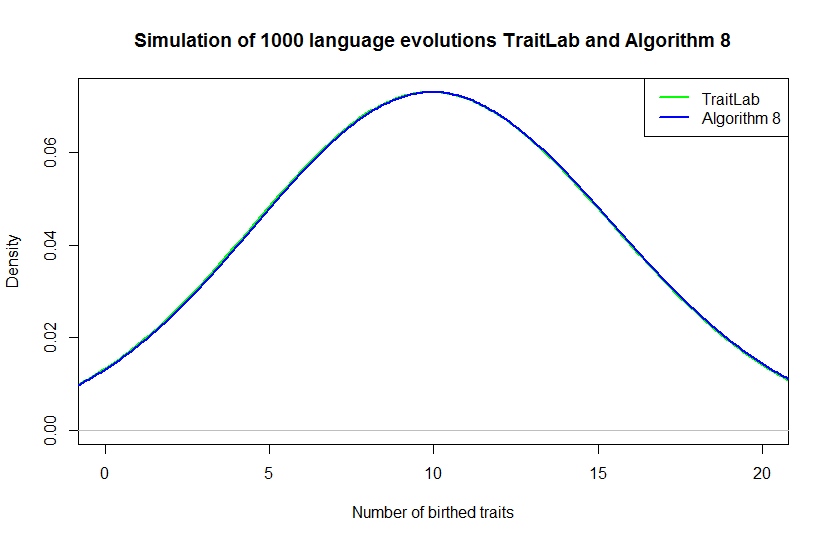}
				\label{fig:TraitLab}
                \caption{Comparison of TraitLab \citep{Nicholls-2011}, and Algorithm \ref{alg:whole_tree_gen_sd}.}
			\end{figure}
            Using the default trait loss rate ($0.2$) and borrowing rate ($0.1$) in TraitLab, reducing taxa to $8$, and mean traits to $10$, it 
becomes tractable to compare the two programs. To run TraitLab in batch mode, small modifications were made to the public source code 
by one of the authors (David Welch). 
            Figure 7 demonstrates that both simulations produce the same distribution of languages. The number of runs is limited--
comparative to previous validations--due to a limit on the number of runs TraitLab is able to do (1000).  
    
    \newpage
    \section{No Empty Meaning Class}
    
    To keep synthesized data consistent with its real world counterpart, it is important to recognize that each language is unlikely to have no 
cognate for a specific meaning. This is consistent with well known word lists like the Swadesh word list, where every language has a 
cognate for each meaning in the list; due to the universality of the chosen words. 
    
    For this to take place, the cognate substitution models need to be modified to handle the death of a cognate in a more intelligent 
manner. 
    \begin{algorithm}[htbp]
        \caption{No Empty Meaning Class Algorithm}
        \label{alg:no_empty_meaning_class}
        \begin{algorithmic}[1]
            \Function{deathCheck}{$l$} \Comment {Check whether a death event will reduce traits in $l$ to $0$} 
                \If {$l.count(1) > 1$} \Comment {Check number of alive traits}
                    \State \textbf{return} $TRUE$ \Comment{No Empty Meaning Class isn't violated}
                \Else 
                    \State \textbf{return} $FALSE$ \Comment{No Empty Meaning Class is violated}
                \EndIf
            \EndFunction
        \end{algorithmic}
	\end{algorithm}
    
    Algorithm \ref{alg:no_empty_meaning_class} is then run whenever a cognate death event occurs in prior algorithms, determining 
whether or not the death event is legal. 
    
    There may be instances where this behavior is not required, therefore it is coded as an optional flag. 
    	
    \section{Missing Data}
    
    Realistically, linguistic data is often incomplete. In the case of ancient languages the passage of history often obscures or destroys 
records, and through this marginally adequate data, gaps in the lexicon form. The \textit{Comparative Indo-European Database} is one 
such example of a database with considerable missing data \cite{Dyen-1992}. 
    
    When producing linguistic models, it is important to determine whether or not missing data is particularly impactful; in some cases 
encoding missing data as absent cognates produces only negligible effects on the outcome of a model \cite{Atkinson-2005}. However, 
where missing data is more prevalent the result of such data manipulation could well be significant. This suggests that models should be 
validated against various types of missing data, as not to introduce biases. 
    
    Where missing data is truly random, it is still possible for that data to be representative of its population. However, where data is 
systematically missing in a non-random way, it can skew analysis of the data in a particular direction. For example, if half of the languages 
in a dataset are missing a particular cognate due to researcher error, analysis will reveal a cognate loss event somewhere in the ancestry, 
where no actual event exists.   
	
    \subsection{Missing Languages}
    
    When a language loses its last native speaker, it becomes a \textit{dead language}, from this point forward any words that have not 
been recorded are lost. Modern languages are much less likely to dodge at least partial recording due to large efforts in the linguistic 
community \cite{Crystal-2000}. Ancient languages are often not as safe. Hunnic is one such language where the sole evidence is a small 
handful of words and names \cite{Pritsak-1982}. Languages like Hunnic can still be used in a phylogenetic approach, but only if models 
are built to handle vast swathes of missing information. 
    \begin{equation}
		\label{eq:missing_lang}
		C = \left[\begin{array}{cccccc} 0 & 1 & 1 & 0 & 1 & 1 \\ ? & ? & 1 & ? & ? & ? \\ 1 & 1 & 1 & 0 & 0 & 1 \end{array}\right]
	\end{equation}
    Equation \ref{eq:missing_lang} demonstrates how a language might be included in a dataset $C$, absent of significant cognate 
information. It still may have useful metadata attached that aids in future analysis.  
    \begin{algorithm}[H]
        \caption{Missing Languages Algorithm}
        \label{alg:missing_lang}
        \begin{algorithmic}[1]
            \Function{missingLanguages}{$A$, $p$} \Comment {Make languages unknown in alignment $a$ unknown with probability $p$}
            	\For {$l$ in $A$} \Comment {For each language in $a$}
                	\State $u \sim B(|l|,p)$ \Comment{Get binomially distributed number of events}
                	\For {$0$..1..$u$}
                    	\State $l[r \sim U(0,|l|)]\:=\:?$ \Comment{Set random cognate to missing}
                    \EndFor
                \EndFor
            \EndFunction
        \end{algorithmic}
	\end{algorithm}
    
    From a technical perspective, the simplest way to randomly simulate missing languages is to do so in a binomial manner given some 
probability $p$, as demonstrated by Algorithm \ref{alg:missing_lang}. 
    
    \newpage
    \subsection{Missing Meaning Classes}
    
    A more common type of missing data is a missing meaning class. While word lists endeavor to be constructed from universal cognates, 
many cognates in these lists are not completely universal \cite{Holman-2008}. Furthermore, for older languages, cognate information may 
be incomplete, leading to situations where models must remain accurate with fewer cognate classes that initially believed.
    \begin{equation}
		\label{eq:missing_mc}
		C = \left[\begin{array}{cc|cc|cc} 0 & 1 & ? & 0 & 1 & 1 \\ 0 & 1 & 1 & ? & 1 & 1 \\ 1 & 1 & 1 & ? & 0 & 1 \end{array}\right]
	\end{equation}
    If Equation \ref{eq:missing_mc} is composed of three meaning classes of two cognates a piece, then its representative of data that is 
missing in the second meaning class. Algorithm \ref{alg:missing_mc} defines a similar binomial method, as seen in Algorithm 
\ref{alg:missing_lang}. 
    \begin{algorithm}[H]
        \caption{Missing Meaning Classes Algorithm}
        \label{alg:missing_mc}
        \begin{algorithmic}[1]
            \Function{missingMeaningClasses}{$MC$, $A$, $p$} \Comment {Make meaning classes $MC$ in alignment $A$ unknown with 
probability $p$}
            	\For {$mc$ in $MC$} \Comment {For each meaning class in $MC$}
                	\State $u \sim B(|l|,p)$ \Comment{Get binomially distributed number of events}
                    \For {$0$..1..$u$}
                    	\State $l = rand(A)$ \Comment{Get random language from $A$}
                    	\State $l[r \sim U(MC)]\:=\:?$ \Comment{Set random cognate in meaning class to missing}
                    \EndFor
                \EndFor
            \EndFunction
        \end{algorithmic}
	\end{algorithm}
        
    \subsection{Algorithm Validation}
    \begin{figure}[H]
				\centering
				\includegraphics[width=12cm]{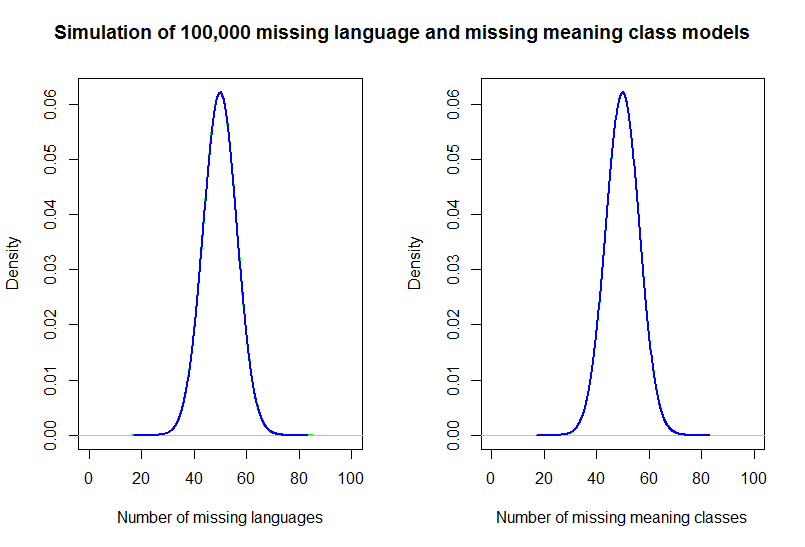}
				\caption{Comparisons of Algorithms \ref{alg:missing_lang} (left) and \ref{alg:missing_mc} (right) in blue, with their 
underlying binomial distributions in green.}
				\label{fig:missing_data}
	\end{figure}
    To validate Algorithms \ref{alg:missing_lang} and \ref{alg:missing_mc}, a 10 language alignment was generated, with 10 meaning 
classes. The algorithms were then run 100,000 with $p=0.5$, and the resulting number of missing languages or meaning classes was 
plotted against the binomial distribution where $n = 10$ and $p = 0.5$.
    
    Figure \ref{fig:missing_data} demonstrates that both algorithms perform as expected, by producing the correct underlying distributions. 
    
    \section{Computational Optimizations}
    
    While the algorithms described in previous sections are functionally correct, the borrowing algorithms (Algorithms 
\ref{alg:whole_tree_gen_gtr} and \ref{alg:whole_tree_gen_sd}) had to undergo additional optimizations to run in satisfactory time. 
    
    \paragraph{Sequence Representation} Initially, sequences were represented as BEAST2 \code{sequence} objects, this allowed for 
easier interfacing with a number of BEAST2 methods that were used throughout the code. The problem with this is the algorithms perform 
a large amount of string manipulation, and each time a language was changed, the \code{sequence} object was unwrapped, modified, 
and wrapped up again. Obviously, this produced significant performance hits in the face of large amounts of manipulation. To remedy this, 
the sequence objects were only updated at branching events, and internal primitive \code{String} representations were kept in the interim. 
    
    It is possible to increase the speedup further here, Java \code{String} objects are not ideal for manipulation. The best solution if a 
language is of constant length is to use a byte array. However, in the case of stochastic-Dollo model, the increasing length of the language 
results in a large amount of time spent copying $n$ size arrays to $n+1$ arrays. While this would have worked for the GTR model, code 
consistency was deemed more important, and speedups were found elsewhere. 
    
    Java's \code{StringBuilder} class was also considered, but ultimately was not implemented because it could not be determined if 
moving from \code{String} to \code{StringBuilder} would produce a significant gain. 
    
    \paragraph{Cognate Memorization} After moving to a faster internal representation of languages, the next bottleneck was the 
calculation of cognate locations. 
    To do most major operations (such as determining the rate of evolution), the number of alive cognates is required. This process 
happens in $O(n)$ time, and usually has to happen for every alive language at any singular point in time. These operations quickly add up 
to substantial computing time, and therefore pave the way for much needed memorization.
    
    Instead of calculating this number on the fly, a running counter can be kept for every language, that is incrementally updated as 
operations on a language occur. While this does increase the memory footprint of the program, it does so with a massive increase in 
speed, and is certainly worth the trade-off. 
    
    \paragraph{Finding an Alive Cognate} After optimizing the representations of the data, the next step is to find functions that are 
particularly time consuming. The most time consuming method--outside of the main loop--is the one to pick a random alive cognate. While 
the number of births have been memorized, their positions have not, leading to another computationally intensive function.
    
    There are two ways to find a random index: deterministically, and non-deterministically. Under the deterministic method, the language 
indices are shuffled randomly ($O(n)$), and this shuffled list of indices is processed until an alive cognate is found ($O(n)$). The resulting 
method has a maximum running time of $O(2n)$, and can be beaten by the non-deterministic method under most cases.
    
    The non-deterministic method simply randomly draws an indices from the language until it finds an alive cognate. The issue here is that 
the method is non-bounded and not guaranteed to complete. The other issue with this method is that under sparse structures (such as 
those languages created using the scholastic-Dollo model), it can perform much worse than the deterministic method. However, the non-
deterministic does perform slightly better for the GTR algorithms. 
    
    \newpage
    To deal with the stochastic-Dollo side of the problem, memorization can once again be employed. When the number of births are 
captured, so can the positions, which can be incrementally updated as operations occur. However, the problem with this is that if Java 
\code{List} structures are employed, removal of indices upon death is $O(n)$, thus moving the intensive operation. The \code{Set} object 
poses a similar problem, in that randomly finding an element in a set takes $O(n)$ time; the problem here is that Java does not allow 
direct access to buckets, unlike C++. 
    
    To gain the speedup required, a custom set class can be defined which avoids linear time operations, but this has not currently been 
implemented. 
    
    \section{Model Misspecification}
    
    One use of simulated data is to determine the accuracy of a models estimates under different scenarios. In the case of linguistics, 
some critics suggest the application of biological phylogenetics to languages and cultures isn't completely apt. It is suggested that certain 
language phenomena (such as word borrowing), invalidate phylogenetic assumptions; in turn, producing spurious results \citep{Moore-1994}, and \citep{Terrell-1988}. 
    
    \subsection{Testing for Misspecification}
    
    When simulating languages, it is done down an evolutionary tree. The leaf languages of this tree can then be fed into phylogenetic inference packages based on \textit{Markov chain Monte Carlo (MCMC)} methods to recover the original tree. The difference between the original and the tree --- estimated via MCMC --- is then indicative of how sensitive the inference model is to misspecification.

    Given a the tree-space of all possible trees, MCMC algorithms in general randomly move through this space, proposing new trees 
based on the current tree then accepting and rejecting moves from one tree to another; based on the probability of the tree under the 
model. This process is repeated millions of times --- with trees in early generations being discarded (\textit{burn-in}) --- until the MCMC 
algorithm reaches a stationary distribution.
    
    Once a sufficient number of trees have been sampled by an MCMC algorithm, they must be compared to the original tree to determine 
the amount of misspecification. There are a number of distance metrics for such a comparison, that are discussed below.
    
    \subsection{Misspecification Due To Borrowing}
    
    The effect of word borrowing on model misspecification has been studied previously; 	\citet{Greenhil-2010} looks specifically at the 
stochastic-Dollo model of language evolution and concludes that under realistic levels of borrowing, phylogenetic inference remains 
robust. This conclusion is consistent for both global and local borrowing. The experiment itself was performed using TraitLab, and 
simulating data via BEAST2 allows for both replication of the stochastic-Dollo results, and further testing under the GTR model. 
    
    \citep{Currie-2010} modeled borrowing of cultural traits as opposed to individual words. This was done using a Brownian motion model, 
with the additional complication of the correlation of traits. While not an entirely similar experiment, results showed that phylogenetic 
inference was once again robust to realistic levels of borrowing. 
    
    One study that suggests that borrowing may cause problems is \citep[pp. 111-118]{MacMohn-2005}. In this study, borrowing levels 
about $10\%$ produced trees very different from the original. There are two possible reasons for this: firstly, borrowing only occurred on 
leaf nodes in the final generations, and secondly, they use less powerful distance measures to reconstruct trees. What this study does 
highlight however, is the necessity for studying borrowing under different experimental conditions. As well as the need for further 
replication where possible. 
    
    \subsubsection{Material and Methods}

	To determine the effect of borrowing on the effect of phylogenetic estimates, languages are simulated under both the GTR, and 
stochastic-Dollo models of language evolution. Once a set of languages is produced (from the leaves of an input tree), BEAST2's MCMC 
algorithms are used to rebuild the most probable trees, which are then compared to the original input tree. 
    
    \newpage
    \paragraph{Root Languages for Sequence Generation} At the root of the tree, the input language needs to represent realistic data. In the format defined in Equation \ref{eq:lang_example}, a root language would look like a list of alive cognates. In the specific case of the Indo-
European dataset, this list is 2449 cognates in length. This root language is then subdivided into meaning classes equal to its length, 
representing a single trait per meaning class. 
    \begin{figure}[H]
				\centering
				\includegraphics[width=10cm]{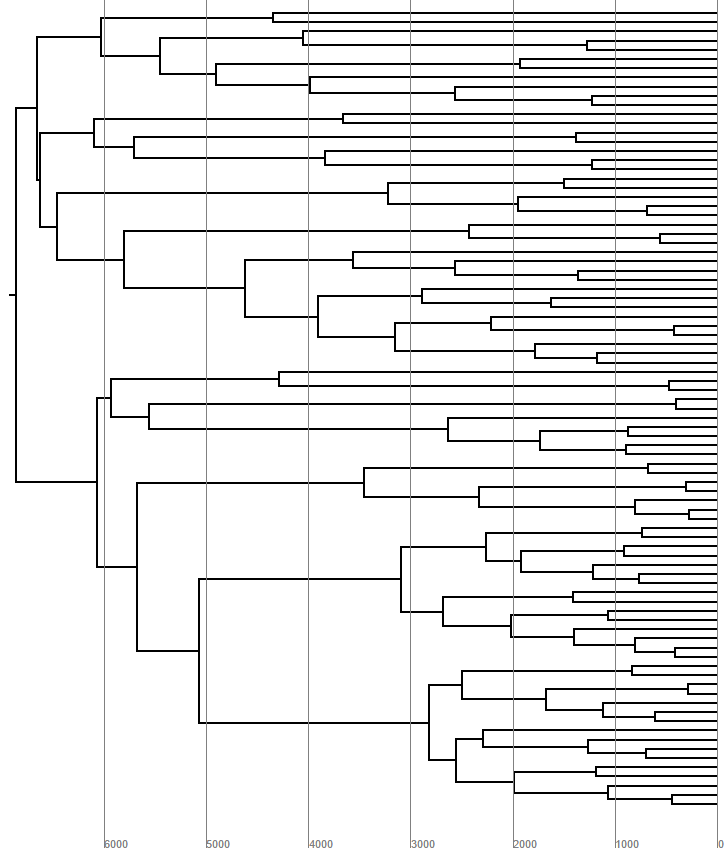}
				\caption{Tree reconstruction of Indo-European dataset for use in creating synthetic data. The tree has a root age of 
6858 years.}
				\label{fig:borrow_tree}
	\end{figure}
    
    \paragraph{Trees for Sequence Generation} The tree itself needs to be representative of realistic data. Using the Indo-European dataset \citep{Dyen-1992}, BEAST2 MCMC analysis was run. 
The result of this analysis was a sampling of probable trees, one of which was randomly chosen as the basis for the average height of generated trees. The trees themselves are randomly generated on each run, with 80 leaves, and $\lambda = 0.00055$. The tree used for the height basis can be seen in Figure \ref{fig:borrow_tree}.     
    
    \paragraph{Borrowing Rates}  To determine what rates of borrowing should be used, a relationship can be defined between the 
percentage of traits borrowed over a $1000$ years, and the rate of borrowing $b$. 
    \begin{equation}
		\label{eq:borrow_percent}
		Borrowing\:Percentage = 1 - \exp(-b1000)
	\end{equation}
	\begin{table}[H]
		\centering
		\caption{Borrowing rates and associated percentages defined by Equation \ref{eq:borrow_percent} for the GTR model.}
		\label{tab:borrowing_rates_percentages}
		\resizebox{\textwidth}{!}{%
		\begin{tabular}{l|lllllllll}
			\textbf{Rate}       & 0 & 0.045 & 0.224 & 0.448 & 0.672 & 0.896 & 1.344 & 1.793 & 2.241 \\
			\textbf{Percentage} & 0 & 1     & 5     & 10    & 15    & 20    & 30    & 40    & 50   
		\end{tabular}
		}
	\end{table}
    
    \paragraph{MCMC Run Parameters} For each model, 100 MCMC runs were performed using a chain length of 22,000,000 of which 10\% was discarded as burnin. However, the covarion model was run for double this length (44,000,000), as certain parameters did not converge satisfactorily 
in 22,000,000 steps. This resulted in 2,000 trees per run.
    
    For all models a strict clock and yule tree priors were used. For the substitution models, GTR inference runs requires the morph-model package \citep{Morph-2014}. For Covarion inference, the Babel package was used \citep{Babel-2015}. Both these packages supplied additional models not available in native BEAST2. 
    
    \paragraph{Tree Comparison} In \citep{Greenhil-2010}, the quartet distance measure is used \citep{Day-1986}. This measure defines the difference between two trees as the number of differing subsets of four leaves that do not have related topology. While this is a good measure, it does not take into account other tree parameters: such as tree height. 
    
    The implementation of the quartet distance measurement used is tqDist \citep{Sand-2013}. While this can't handle the output of BEAST2 directly, only small modifications are required. This distance is measured between the input tree and each inferred tree, and multiple samples per run are captured to deal with perturbations. 
   
    Height is measured as the difference between the randomly generated input tree. To account for variable starting tree heights, differences were normalised to a value between 0 and 1. 
    
    Both measures are done for a random sample of 300 trees in each MCMC run, this sample is then averaged to produce a single value per run, resulting in 100 data points per borrowing rate. In the figures below, these are plotted along with a mean for the entire rate, and the 94\% highest posterior density (HPD) interval.
    
    \newpage
    \subsubsection{GTR Model}
    For the GTR model, the mutation rate $\mu$ is set such that the rate at which the probability of moving between states 1 to state 0 (or vice-versa) over a 1000 year period is 0.1, which is representative of the rate of change in the Indo-European dataset.
    \begin{figure}[H]
		\centering
		\includegraphics[width=12cm]{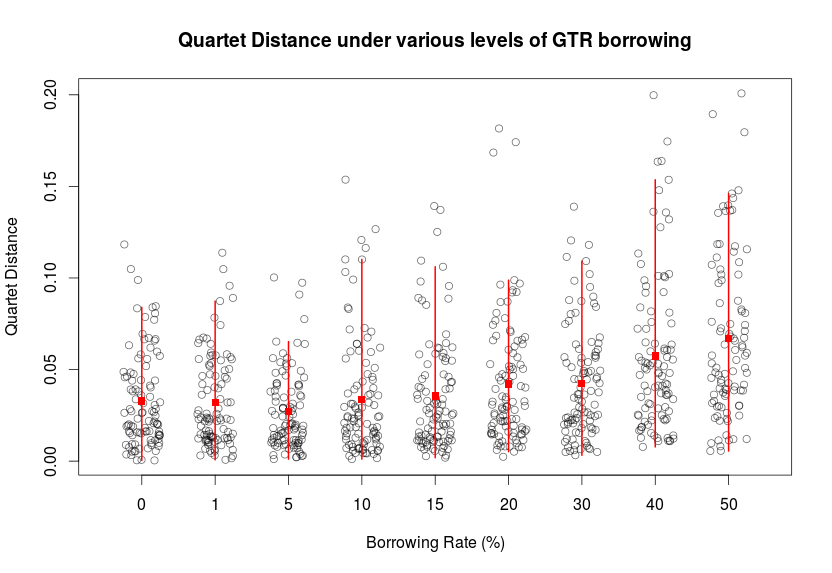}
		\caption{Quartet distances of inferred trees using synthetic languages created under the GTR model. The red square indicates the mean, and the red line indicates the 95\% HPD interval.}
		\label{fig:quart_gtr}
	\end{figure}
    When GTR languages are inferred under the GTR model, the mean quartet distance (Figure \ref{fig:quart_gtr}) does not deviate from the baseline ($0.033$), until reasonably high levels of borrowing ($40\%>$). At 50\% borrowing, the mean distance increases to $0.067$.
    
    The HPD interval begins to widen at lower levels of borrowing. Beginning at $[0.0,0.102]$, the interval begins to increase for rates of 10\% and above. At 10\%, the interval increases to $[0.0,0.115]$, and continues to increase; at 50\% the interval is: $[0.0,0.165]$.
    \begin{figure}[H]
		\centering
		\includegraphics[width=12cm]{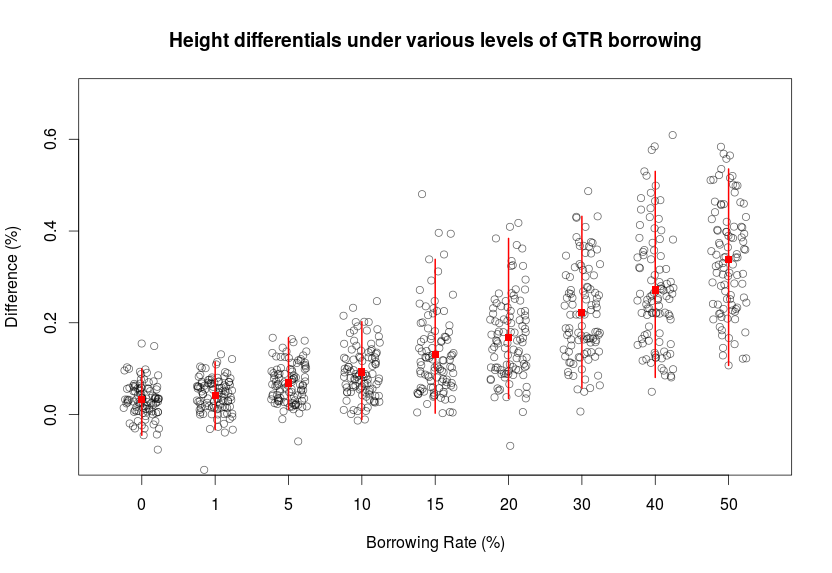}
		\caption{Percentage differences in heights between the original tree used for language generation, and the inferred trees under the GTR model. The red square indicates the mean, and the red line indicates the 95\% HPD interval.}
		\label{fig:height_gtr}
	\end{figure}
    Differences between true tree height, and inferred tree height trend upward as the borrowing rate increases. Figure \ref{fig:height_gtr} shows that the mean height moves from $0.033$ at 0\%, to $0.338$ at 50\%. This trend begins earlier when compared to the GTR quartet distances, with the mean perceptibly increasing at the 5\% level ($0.068$).
    
    The HPD interval also increases with borrowing rate, starting at $[-0.106,0.162]$ at 0\% borrowing, and trending upwards to $[-0.074,0.206]$ at 50\%.
    
    Figure \ref{fig:height_gtr} also highlights that a number of runs produced what could be considered outlying values. 
    
    \subsection{Stochastic-Dollo Model}
    
    For the simulations, the death rate $\mu$ is defined by the mean proportion of traits lost per 1000 years. This was kept the same as the GTR model, at $l=0.1$.
    \begin{equation}
		\label{eq:death_rate_borrow}
		\mu = \frac{-\log(1-l)}{1000}
	\end{equation}
   	To determine the birth rate $\lambda$, the death rate is multiplied by the length--$L$--of the root language.
    \begin{equation}
		\label{eq:birth_rate_borrow}
		\lambda = L\mu
	\end{equation} 
    To model the stochastic-Dollo synthetic languages, two models were picked due to their common use in inference processes: GTR and Covarion. 
    
    \subsubsection{GTR Inference}
    \begin{figure}[H]
				\centering
				\includegraphics[width=12cm]{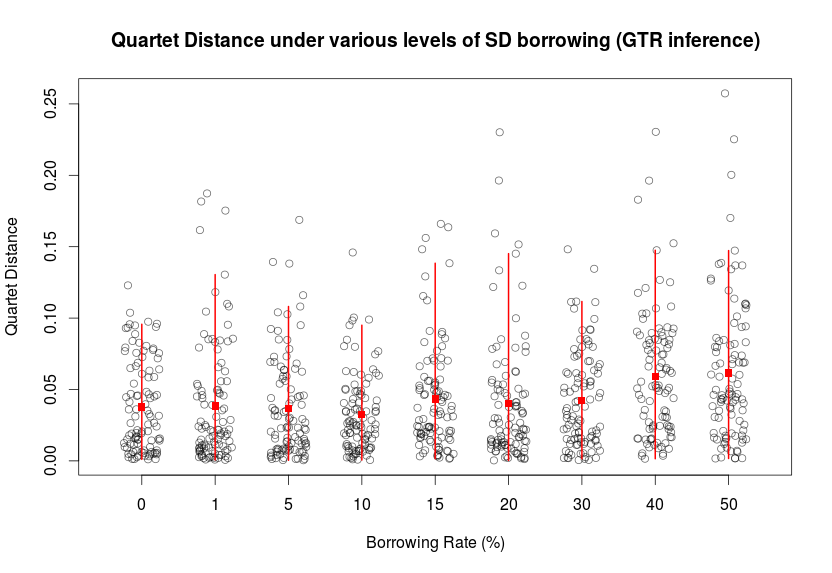}
				\caption{Quartet distances of inferred trees using synthetic languages created under the SD model, and inferred using the GTR model. The red square indicates the mean, and the red line indicates the 95\% HPD interval.}
				\label{fig:quart_sd}
	\end{figure}
	When synthesizing languages under the SD model, and inferring topology under the GTR model, quartet distance increases in a similar pattern to GTR synthesis: a perceptible effect is only noticed at the highest levels of borrowing.
    
    Deviation from the baseline ($0.038$) does not occur until 40\% borrowing, at which point the mean increases to $0.059$. 
    
    The change in HPD intervals is somewhat less clear, as the 20\% interval ($[0.0,0.149]$) is similar to the 40\% interval ($[0.0,0.147]$). 
    \begin{figure}[H]
		\centering
		\includegraphics[width=12cm]{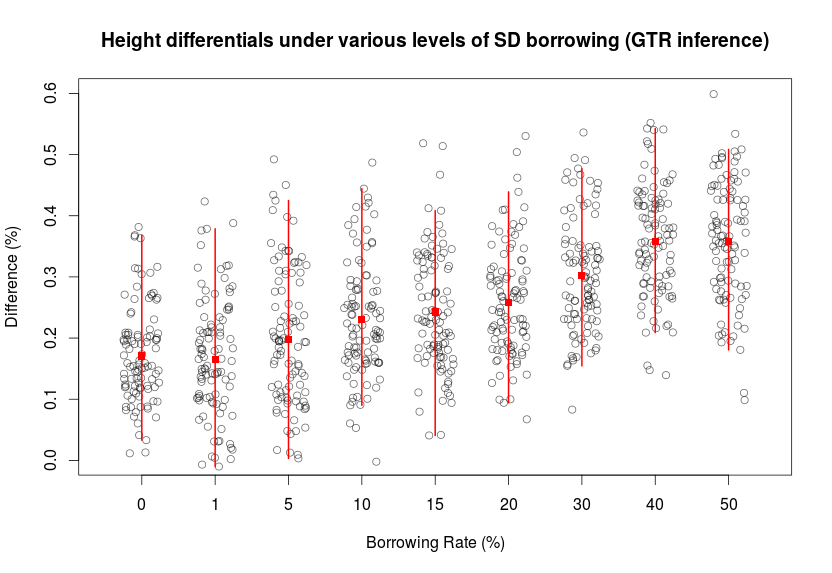}
		\caption{Percentage differences in heights between the original tree used for language generation, and the inferred trees under the SD model, and inferred under the GTR model. The red square indicates the mean, and the red line indicates the 95\% HPD interval.}
		\label{fig:height_sd}
	\end{figure}
    Figure \ref{fig:height_sd} shows a strong relationship between the mean height difference and the borrowing rate. As the rate increases, the mean difference moves upwards away from the baseline ($0.170$) to $0.358$ at 50\%.
    
    The HPD interval however, does not increase significantly. At 0\% borrowing, the interval is $[-0.010,0.372]$; at 50\%, the interval is [0.160,0.541]. The difference between these two intervals is $0.1$. Although, it is worth noting the upward movement of the 50\% interval. 
    
    \subsubsection{Covarion Inference}
    \begin{figure}[H]
		\centering
		\includegraphics[width=12cm]{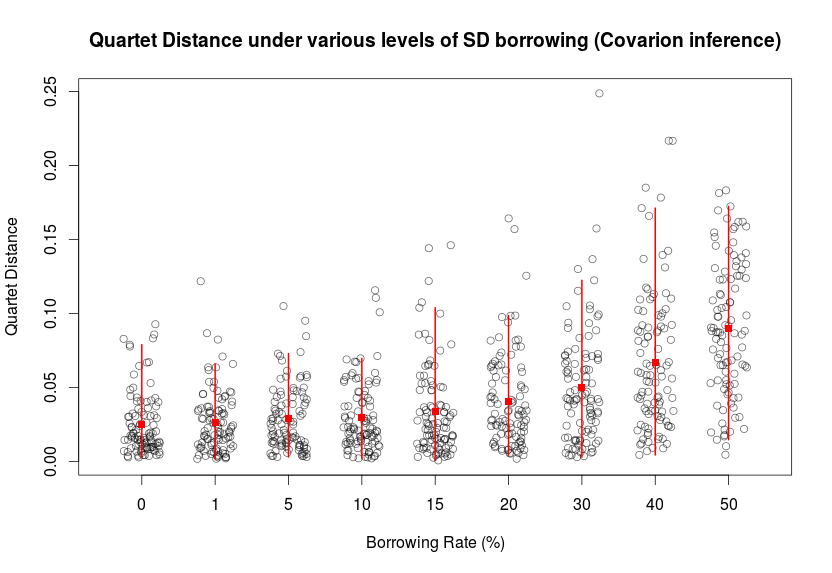}
		\caption{Quartet distances of inferred trees using synthetic languages created under the SD model, and inferred using the Covarion model. The red square indicates the mean, and the red line indicates the 95\% HPD interval.}
		\label{fig:quart_cov}
	\end{figure}
	When data is inferred under the Covarion model, there is a stronger relationship between quartet distance and borrowing rate when compared to previous experiments. 
    
   In Figure \ref{fig:quart_cov}, movement away from the mean baseline ($0.025$) begins at 15\% borrowing ($0.034$), and continues to increase with the borrowing rate. At 50\% borrowing, the mean quartet distance is $0.089$.
    
    This same pattern is reflected in the HPD intervals, with the baseline starting at $[0.000,0.091]$, and 50\% having an interval of $[0.000,0.178]$.
    \begin{figure}[H]
		\centering
		\includegraphics[width=12cm]{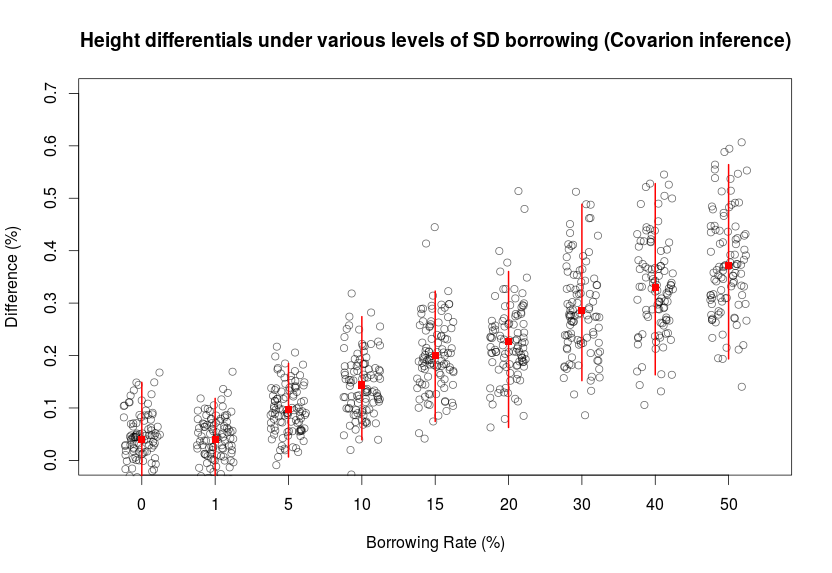}
		\caption{Percentage differences in heights between the original tree used for language generation, and the inferred trees under the SD model, and inferred under the Covarion model. The red square indicates the mean, and the red line indicates the 95\% HPD interval.}
		\label{fig:height_cov}
	\end{figure}
	Once again, the correlation between height differential and borrowing rate appears to be quite strong in Figure \ref{fig:height_cov}. From the mean baseline of $0.040$, increases begin at 5\% borrowing ($0.097$), and continue to 50\% ($0.372$). 
    
    Again, this pattern is reflected in the HPD intervals. The baseline interval begins at $[-0.122,0.197]$, and by 50\% borrowing, has increased to $[0.187,0.593]$.
    \subsection{Discussion}
    
    In all experiments, there is a clear correlation between an increasing borrowing rate, and further deviation from the true tree. 
    
    Before discussing the magnitude of these effects, it is important to provide context for both measurements. For quartet distance, any two random yule trees (with $80$ leaves) have a mean quartet distance of $0.667$, so as inferred trees approach this value, they essentially become no better than randomly assigning leaf topologies.
    
    For the height measurement, positive differences between the true tree and inferred tree heights is symptomatic of underestimation. For example, if the true tree is $7000$ years old, then a difference of $0.2$, means that the inferred tree has an age of $5600$ years. What this shows, is that even small height differences can actually translate into huge miscalculations.

    \subsubsection{Topology}
    
    For quartet distance, the relationship under both the GTR and the stochastic-Dollo model is one that doesn't take serious effect until very high levels of borrowing. All three simulations show a similar range of quartet distances ($[0,0.25]$), and both the GTR and SD (Covarion inferred) simulations show an increasing mean quartet distance from 15\% borrowing. However, increased variability in the SD (GTR inferred) data means that this relationship is less clear until around 40\%. 
    
    The SD (GTR inferred) distances have a different relationship when compared to the SD (Covarion inferred) distances. In Figure \ref{fig:quart_sd}, the relationship is relatively unclear until about 40\% borrowing. However, Figure \ref{fig:quart_cov} shows a  much clearer relationship beginning as early as 15\%. This is because while the Covarion model begins with a lower mean ($0.025$ versus $0.038$), by the end of the graph, its mean is in fact higher ($0.089$ versus $0.067$). What this means is that SD data is better inferred under the Covarion model, as there are fewer outliers in the Covarion case.
    
    What is important to remember that while there are obvious relationships, the mean distances are very small for all levels of borrowing. Where inference is only concerned with tree topology, reasonable borrowing appears to have a minimal effect. 
    
    This isn't to say that there isn't cause for concern at high levels of borrowing. Both the GTR and the SD have multiple outliers significantly higher than the baseline of 0\% borrowing. While GTR only reaches 0.311, the SD model (inferred under both the GTR and Covarion models) has outliers as high as 0.686, which is above what is considered a random quartet distance. 

    \subsubsection{Tree Height}
    
    When attempting to infer heights under levels of borrowing, the effects are much more pronounced. Under the GTR simulation, a steady increase to a mean difference of $0.338$ (50\%) results in the average age of a 7000 year old tree being underestimated by 2336 years. This effect is intensified by a steadily increasing HPD interval, which highlights the uncertainty of measurements.
    
    Even at smaller (15\%) borrowing levels, a 0.131 mean difference results in underestimation by $917$ years. What is also noticeable, is that even at this level the range is very large, with outliers as far out as $0.762$ (underestimation of $5334$ years). Both the GTR and SD (Covarion inferred) simulations show clear trends of increasing underestimation as the borrowing rate increases.
    
    The relationship is less pronounced in the SD model (GTR inferred), but only because the baseline mean is already significantly underestimating true tree height by $0.167$. At the highest levels of borrowing, the mean and outlying estimations are similar to the GTR model, but the the 15\% level has a markedly higher mean of $0.236$. Increased scatter in the SD (GTR inferred) simulation also highlights that Covarion is a better inference method, as stated in the previous section. 
    
    When performing inference under the Covarion model we see a strong relationship between borrowing level and inferred tree height. Once again, this due to the fact that the Covarion baseline mean begins lower than the SD (GTR inferred) mean ($0.04$ versus $0.170$). However, as borrowing increases the difference between the two inference methods decreases, and both end up with a similar mean at the 50\% level ($0.372$ versus $0.358$). 
    
    Ultimately, trying to infer tree height under some level of borrowing is going to result in the underestimation of the date, increasing in severity as the borrowing rate increases. 
    
    \section{Conclusion} 
    
    Due to extensive work already completed by the group behind BEAST2, extending it to synthesize languages is a more surmountable--but still difficult--challenge. Since one of the primary drivers of generating synthetic data is model validation, being able to generate and then experiment on models in a single system is a real advantage over previous methods. 
    
    While the basis for language generation models has been created in this thesis, it has been done in such a way that new models of generation can be implemented as needed by researchers, without having to write entire new plugins for BEAST2, or new external systems. 
    
    BEAST2 did have sequence generation already built in. However, this generation was aimed at biological data, and did not have features specific to linguistic models, such as borrowing and missing cognate models. These models are once again built in such a way that they can be extended and altered in ways that any future analyses may demand. 
    
    Finally, as a proof of work, the effects of linguistic borrowing on inference were studied. The results of this analysis not only demonstrate the effectiveness of the tools described previously, but highlight the borrowing effect on both the topology and heights of inferred trees. As the borrowing rate increases, both quartet distance and height difference increase, resulting in lower accuracy of inferred trees. Understanding and accounting for these borrowing effects will allow for future work to produce more accurate descriptions of language relationships. 
    
    \newpage
	\bibliography{thesis_bib}
	\bibliographystyle{abbrvnat} 
    
    \newpage
    \appendix
    
    \section{Language Sequence Generation Plugin Guide}
    
    The plugin described in this thesis can be found at: \url{https://github.com/lutrasdebtra/Beast-Borrowing-Plugin}.
    
    \subsection{Quick Start}
    
    The project can easily be cloned as a normal java project, however, it does require the BEAST2 \citep{Bouckaert-2014} project in it's 
build path, a copy of which is included in the \code{lib} folder. Alternately, it can run directly as a \code{jar}, either through compilation, or 
from \code{Jars/LangSeqGen.jar}.
    
    \subsubsection{Command Line}
    
    Like the original BEAST2 \code{seqgen}, this plugin uses the same format for command line runs:
\begin{lstlisting}[language=Java]
java LanguageSequenceGen <beast file> <nr of instantiations> [<output file>]
\end{lstlisting}
    \begin{itemize}
		\item The \code{<beast file>} is an \code{xml} file that specifics the initial input parameters. An example is provided below.
        \item To determine the number of meaning classes, \code{<nr of instantiations>} is provided, the position of first cognate in each 
meaning class is provided as an additional sequence at the end.
        \item If an \code{<output file>} is not provided, the output will be written to \code{std.out}.
	\end{itemize}
    
    \subsubsection{BeastBorrowingPluginTest}
    
    Like most BEAST2 plugins, this plugin has its own testing suite defined in \code{BeastBorrowingPluginTest}. In this class, the 
\code{SeqGenTest()} runs the plugin using arguments defined within the function:
\begin{lstlisting}[language=Java]
private static void SeqGenTest() {
        String[] args ={"examples/testSeqLangGen.xml",
        "2",
        "examples/output.xml"};
        LanguageSequenceGen.main(args);
}
\end{lstlisting}
The format of the arguments are the same as those in the \textit{Command Line} section.

There are a number of other tests in the class that produce \code{csv} files, which are in turn used to validate various portions of the 
plugin in R.

	\subsection{Explanation of the Input/Output Files}
    
    \subsubsection{The BEAST File}
    
    The BEAST file outlines how to produce the synthetic data. An example is provided in \code{/examples/testSeqLangGen.xml}; it is 
reproduced below:
\begin{lstlisting}[language=XML]
<beast version='2.0'
       namespace='beast.evolution.alignment:
       beast.evolution.substitutionmodel'>

    <tree id='tree' spec='beast.util.TreeParser' IsLabelledNewick='true' newick='((((english:0.02,(german:0.01,
    french:0.01):0.4):0.01,spanish:0.3):0.2,
    italian:0.6):0.3,irish:0.6)'/>


    <run spec="beast.app.seqgen.LanguageSequenceGen" tree='@tree'>
        <root spec='Sequence' value="01010101010100100010101010000100" taxon="root"/>

        <subModel spec='ExplicitBinaryStochasticDollo' birth="0.5" death = "0.5" borrowrate ="0.0" borrowzrate="0.0" noEmptyTrait="false"/>    
    <missingModel spec='MissingLanguageModel' rate="0.5"/>
    </run>
</beast> 
\end{lstlisting}
	\begin{itemize}
		\item The \code{tree} takes a newick formatted tree with both branch distances and taxon node names.
        \item The \code{run} initiates the plugin using the \code{tree} defined above. It also has a number of interior parameters:
        \begin{itemize}
			\item \code{root} is the sequence to be placed at the root of the tree. It should consist of present (1) or absent traits (0). 
The plugin does not handle missing or unknown traits. The \code{taxon} does not need to be \textit{root}.
			\newpage
            \item \code{subModel} defines the model used to simulate evolution down the tree. All models have a \code{borrowrate} 
parameter, which defines the rate of global borrowing; \code{borrowzrate} defines the distance of local borrowing; note: if 
\code{borrowzrate} is set to \code{0.0}, the plugin assumes an infinite distance. Currently there are two models:
            \begin{itemize}
				\item \code{ExplicitBinaryGTR} evolves the \code{root} via a Generalised Time-Reversible model. This model has a 
single \code{rate} parameter which defines the rate at which traits both can be birthed, and die.
                \item \code{ExplicitBinaryStochasticDollo} evolves the \code{root} via a stochastic-Dollo model of sequence evolution, which has 
both a \code{birth} rate of traits, and a separate \code{death} rate.
			\end{itemize}
            \item \code{missingModel} defines the model used to simulate missing data in the final alignment. Currently, this is non-optional 
and to not use it \code{rate} should be set to \code{0}.
            \begin{itemize}
				\item \code{MissingLanguageModel} - Each language has a random binomial number of missing events, which 
convert random cognates in the language to \code{?}.
                \item \code{MissingMeaningClassModel} - Each meaning class has a random binomial number of missing events, which convert 
random cognates in the meaning class to \code{?}.
			\end{itemize}
		\end{itemize}
	\end{itemize}
    
    \newpage
    \subsubsection{The Output File}
    
    The Output file is a simple BEAST2 \code{alignment} piped to \code{xml}. An example from \code{/examples/output.xml} can be found 
below:
\begin{lstlisting}[language=XML]
<beast version='2.0'>
<data id='SD' dataType='binary'>
    <sequence taxon='english' value='111111111111111111111111'/>

    <sequence taxon='german' value='111111111111111111111111'/>

    <sequence taxon='french' value='111111111111111111111111'/>

    <sequence taxon='spanish' value='111111111111111111111111'/>

    <sequence taxon='italian' value='111111111111111111011111'/>

    <sequence taxon='irish' value='111111111111111101101111'/>
</data>

<!-- Meaning Classes: 0 -->
<!-- Created at: 2016-04-26 15:09:16.506 -->
</beast>
\end{lstlisting}

\subsection{The Thesis Analysis Package}

Included in the repository is a number of classes inside \code{beastborrowingplugin/thesisanalysis}. These classes are used to do batch 
analysis of sythetic languages produced by this package under BEAST2 inference.

These classes are not required for the main running of the program, but may be useful if batch analysis is needed.

\end{document}